\documentclass[times,twocolumn,final,author]{elsarticle}

\usepackage{prletters}
\usepackage{framed,multirow}

\usepackage{latexsym}

\usepackage{url}
\usepackage[hidelinks]{hyperref}
\usepackage{xcolor}
\definecolor{newcolor}{rgb}{.8,.349,.1}

\usepackage{placeins}
\usepackage{amsmath,amssymb,amsfonts}
\usepackage{algorithmic}
\usepackage{graphicx}
\usepackage{caption, subcaption, floatrow}
\usepackage{textcomp}
\usepackage{xcolor}
\usepackage{booktabs}

\usepackage{tikz}

\journal{Pattern Recognition Letters}

\begin{document}

\clearpage
\thispagestyle{empty}
\ifpreprint
	\vspace*{-1pc}
\fi

\clearpage
\thispagestyle{empty}

\ifpreprint
	\vspace*{-1pc}
\else
\fi

\clearpage

\ifpreprint
	\setcounter{page}{1}
\else
	\setcounter{page}{1}
\fi

\begin{frontmatter}

	\title{Large Margin Classifier with Graph-based Adaptive Regularization}

	\author[1]{Vítor~M.~Hanriot}
	\ead{vhanriot@ufmg.br}
	\author[1]{Turíbio~T.~Salis}
	\ead{turibiotanussalis@ufmg.br}
	\author[2]{Luiz~C.~B.Torres\corref{cor1}}
	\ead{luiz.torres@ufop.edu.br}
	\author[1]{Frederico~Coelho}
	\ead{fredgfc@ufmg.br}
	\author[1]{Antonio~P.~Braga}
	\ead{apbraga@ufmg.br}
	\cortext[cor1]{Corresponding author}

	\address[1]{Graduate Program in Electrical Engineering - Universidade Federal de Minas Gerais - Av. Antônio Carlos 6627, 31270-901, Belo Horizonte, MG, Brazil}
	\address[2]{Department of Computer and Systems - Universidade Federal de Ouro Preto - João Monlevade 35931-022, Brazil.}

	\begin{abstract}
		This paper introduces the use of per-class regularization hyperparameters in Gabriel graph-based binary classifiers. We demonstrate how the quality index used for regularization behaves both in the margin region and in the presence of outliers, and how incorporating this regularization flexibility can lead to solutions that effectively eliminate outliers while training the classifier. We also show how it can address class imbalance by generating higher and lower thresholds for the majority and minority classes, respectively. Thus, rather than having a single solution based on fixed thresholds, flexible thresholds expand the solution space and can be optimized through hyperparameter tuning algorithms. Friedman test shows that flexible thresholds are capable of improving Gabriel graph-based classifiers.
	\end{abstract}


\end{frontmatter}

\section{Introduction}

Learning from data may require achieving a trade-off between error and model complexity. The problem was formalized in the literature a few decades ago~\citep{geman,vapnik1995nature} and has been treated from different perspectives throughout the years with approaches like model shrinking~\citep{Reed,lecun1989optimal}, regularization~\citep{Girosi,SILVESTRE2015288} and multiobjective learning~\citep{ALBUQUERQUETEIXEIRA2000189}. Essentially, such methods work by limiting model size or the magnitude of parameters, aiming to reduce the effects of overfitting inherent to oversized models trained with error minimization objective functions. Minimizing error often requires higher complexity terms in the approximation function. However, the flexibility introduced by these terms can also lead to the incorporation of data uncertainty into the model. Since most current machine learning models are based on the superposition of functions with the same complexity (e.g. sigmoidal or radial basis functions), these higher-complexity terms are difficult to identify, and the usual approach is to limit the magnitude of their parameters, so regularization and multi-objective learning are the most common approaches.

This paper introduces per-class regularization hyperparameters to Gabriel graph (GG)~\citep{Gabriel_1969}-based classifiers, such as Chipclass~\citep{Torres_2015} and GG-based Gaussian Mixture Models (GMM-GG)~\citep{torres2020large} and Radial Basis Function Networks (RBF-GG)~\citep{torres2014geometrical}. These classifiers are based on the principles of support vectors (SVs), which are obtained through quadratic programming (QP) in Support Vector Machines (SVMs)~\citep{Vapnik_1992}, through solving a set of linear equations by using equality constraints in Least Squares SVMs (LS-SVMs)~\citep{suykens1999least} or through combining concepts from SVMs and LSSVMs by using a two-step verification approach to select SVs in IP-LSSVM~\citep{Carvalho_2009}. In contrast with SVMs, however, these GG-based classifiers rely on structural support vectors~\citep{hanriot2024multiclass}. In Chipclass~\citep{Torres_2015}, for instance, a global separator is composed as the aggregation of local maximum margin hyperplanes: the bisecting hyperplanes defined by all graph edges formed by vertices from opposite classes are the basic elements that compose the global classifier. The method is, therefore, easy to implement, utilizes a computational geometry approach rather than requiring host processors to execute optimization algorithms, is feasible to hardware implementation~\citep{Garcia_2020,janier2022,Torres_2015} and does not demand user intervention to set parameters. In order to avoid overfitting, local hyperplanes are pruned according to graph properties and a pre-established strategy to eliminate those that may be composed by outliers, a common step for GG-based classifiers~\citep{Torres_2015, torres2020large, torres2014geometrical}. The basic principle of this strategy is that outliers are most likely responsible for those higher complexity terms of the aggregated function, so they are identified by a quality index assigned to each pattern according to their graph relations with elements from the opposite class.
Parsimonious models which are statistically equivalent to SVMs selected with cross-validation have been reported~\citep{torres2020large,TORRES2022192} by considering outlier thresholds as the mean values of the quality index of each class. This strategy is in accordance with the principle of providing a stand-alone autonomous classifier that can fully learn from data without user interference.

This paper relaxes the autonomy principle and explores the idea of implementing GG-based classifiers with an adaptive threshold, aiming for the explicit optimization of an objective function through hyperparameter tuning, a strategy that is adopted by most learning models. For instance, SVMs require kernel and regularization parameters to be adapted. The idea in the particular case of GG-based classifiers is to trade-off autonomy of the model by gain in performance, since the model has stand-alone properties when the original strategy is adopted. Thus, the contributions of this paper are:

\begin{itemize}
	\item Addition of per-class regularization hyperparameters in Gabriel graph-based classifiers, allowing selective elimination of outliers while preserving the margin region.
	\item Adaptive thresholding using such hyperparameters can generate higher thresholds for the majority class and lower thresholds for the minority class, improving class balance and discrimination.
	\item With hyperparameter tuning and cross-validation, GG-based models with flexible thresholds present better average ranks than their standard versions and have performance that is statistically equivalent to other state-of-the-art models.
\end{itemize}

Although pre-established thresholds yield good performance models, the results presented in this paper, obtained with threshold selection for both classes via Bayesian Optimization~\citep{snoek2012practical,akiba2019optuna}, show that they can still be improved. Next, section~\ref{sec:rev} provides a brief review of the concepts that are the basis of the proposed method; section~\ref{sec:meth} presents the applied methodology; section~\ref{sec:conc} discusses the proposal; and section~\ref{sec:exp} describes the experiments and results.

\section{Context}
\label{sec:rev}
\subsection{Gabriel graph}

\begin{figure}[h]
	\centering
	\resizebox{3cm}{2.14cm}{
		\begin{tikzpicture}
			[rectin/.style={rectangle,draw=blue,fill=blue,thick,
						inner sep=0pt,minimum size=4mm},
				rectout/.style={rectangle,draw=none,fill=none,
						inner sep=0pt,minimum size=4mm},
				h/.style={circle,color=black,thin,minimum size=10mm},
				texto/.style={}]

			\node[texto] (x1) at (-0.1,0.1) {\tiny $x_1$};
			\node[texto] (x2) at (0.5,1.15) {\tiny $x_2$};
			\node[texto] (x3) at (2.2,0.4) {\tiny $x_3$};
			\node[texto] (x4) at (1.5,1.4) {\tiny $x_4$};
			\node[texto] (x5) at (1.25,-0.48) {\tiny $x_5$};

			\draw[black,fill=blue] (0,0) circle (.2ex);
			\draw[black,fill=blue] (0.5,1) circle (.2ex);
			\draw[black,fill=blue] (2,0.3) circle (.2ex);
			\draw[black,fill=blue] (1.5,1.3) circle (.2ex);
			\draw[black,fill=blue] (1.2,-0.3) circle (.2ex);

			\draw[thick] (0,0) -- (0.5,1) {};
			\draw[dashed,gray] (0,0) -- (2,0.3) {};
			\draw[dashed,gray] (0,0) -- (1.5,1.3) {};
			\draw[thick] (0,0) -- (1.2,-0.3) {};

			\draw[dashed,gray] (0.5,1) -- (2,0.3) {};
			\draw[thick] (0.5,1) -- (1.5,1.3) {};
			\draw[thick] (0.5,1) -- (1.2,-0.3) node[midway,right = -0.1cm] {\tiny $e_{25}$} {};

			\draw[thick] (2,0.3) -- (1.5,1.3);
			\draw[thick] (2,0.3) -- (1.2,-0.3);

			\draw[dashed,gray] (1.5,1.3) -- (1.2,-0.3);

			\draw[black,dashed] (0.85,0.35) circle (4.7ex);

		\end{tikzpicture}
	}
	\caption{Schematic representation of GG construction. Solid lines represent those edges that are included in the graph.}
	\label{GG_main}
\end{figure}
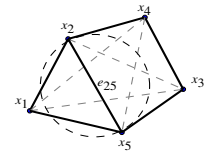

Given a dataset $\mathbf{D}=\{\mathbf{x}_{1}, \mathbf{x}_{2}, \mathbf{x}_{3},\cdots, \mathbf{x}_{m}\}$ where $\textbf{x}_i \in \mathbb{R}^d$,  a \textit{Gabriel graph}~\citep{Gabriel_1969} of $\mathbf{D}$ is an undirected graph of vertices $\mathcal{V} \in \mathbf{D}$ and edges $e_{ij} \in \mathcal{E}$ so that a pair of samples $\{\textbf{x}_i, \textbf{x}_j\}$ is connected  by the edge $e_{ij}$ if, and only if, there are no other elements of $\mathbf{D}$ within the hypersphere having $\textbf{x}_i$ and $\textbf{x}_j$ diametrically opposite. Fig.~\ref{GG_main} shows an example of a GG representation of a dataset with 5 samples $\mathbf{x}_{1}, \mathbf{x}_{2}, \mathbf{x}_{3},\mathbf{x}_{4},\mathbf{x}_{5}$. The figure shows all possible edges for the dataset with solid lines representing those that belong to the graph and dashed lines those that are not included in the final graph representation. The dashed circle with $\mathbf{x}_{2}$ and $\mathbf{x}_{5}$ diametrically opposite shows that there is no other sample within the circle, the reason why edge $e_{25}$ was included.

In practice, the adjacency matrix of a GG is obtained by applying the triangle inequality presented in Eq.~\ref{eq:GabrielEQ},  where $\delta(\cdot,\cdot)$ is a distance operator. Thus, the computation traverses all pairs $(\mathbf{x}_i,\mathbf{x}_k)$ and checks if any other sample falls within the hypersphere defined by $(\mathbf{x}_i,\mathbf{x}_k)$, costing $\mathcal{O}(m^3)$. The resulting graph preserves locality, since edges tend to appear between samples that are spatially close and larger edges tend not to be included. Because of that,  many neighborhood properties of the dataset, relevant for data classification, can be extracted from the graph and considered in data analysis and classification. For instance, edges formed by samples from opposite classes should intercept the separator that minimizes the error. A classifier based on this principle is Chipclass~\citep{Torres_2015}, a method based on a combination of hyperplanes that divide the graph edges located in the separation region.

\begin{equation}\label{eq:GabrielEQ}
	\begin{split}
		e_{ij} \in \mathcal{E} & \leftrightarrow \delta(\mathbf{x}_i,\mathbf{x}_j) \leq \left[\delta(\mathbf{x}_i,\mathbf{x}_k) + \delta(\mathbf{x}_j,\mathbf{x}_k)\right] \: \\
		                       & \forall \: \mathbf{x}_k \in \mathbf{D}, \mathbf{x}_i,\mathbf{x}_j \neq \mathbf{x}_k
	\end{split}
\end{equation}

\subsection{Chipclass}
\label{Chipclass}

\begin{figure}[htbp]
	\centerline{\includegraphics[scale=0.3]{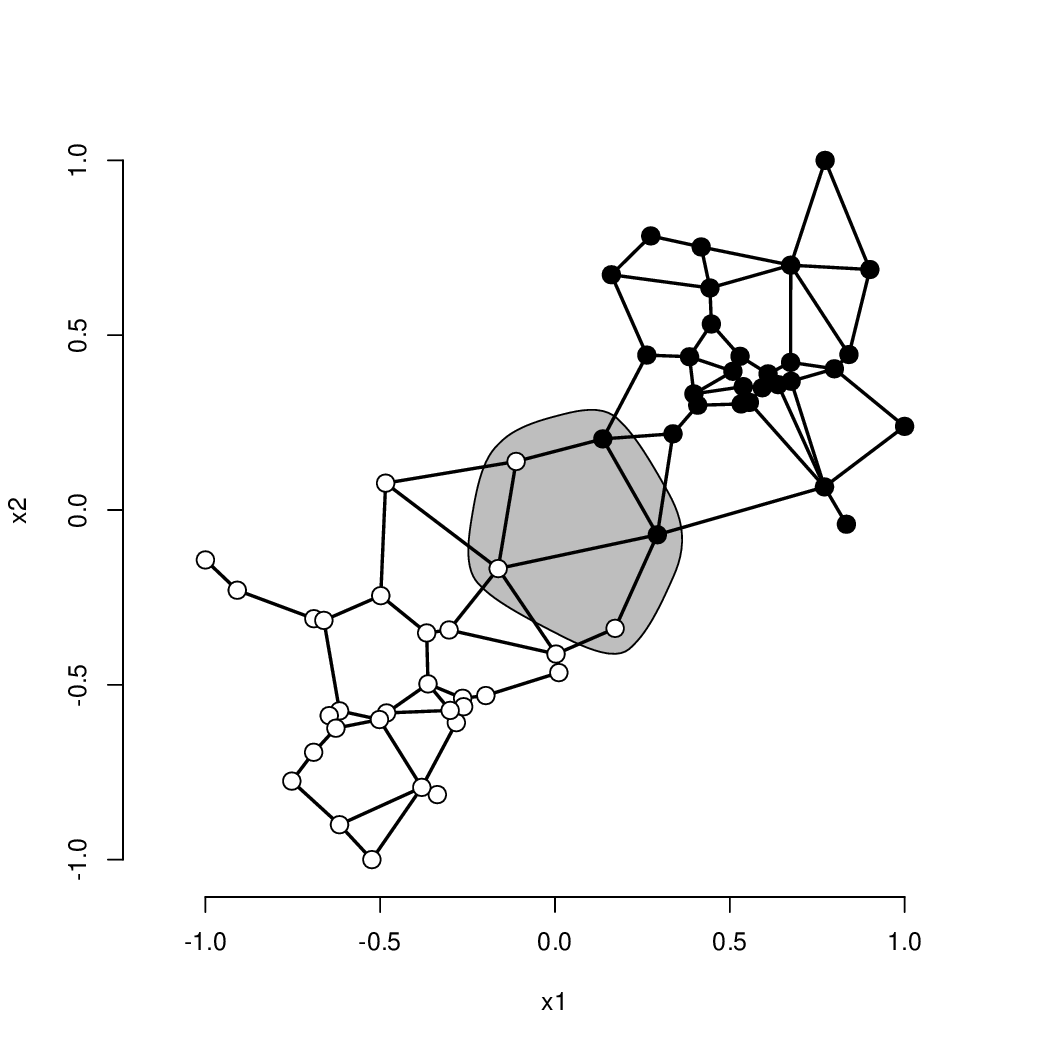}}
	\caption{Example of a binary classification dataset and its corresponding GG. Highlighted SEs form a combined classifier}
	\label{GGClass}
\end{figure}

Chipclass~\citep{Torres_2015} is a large margin classifier derived directly from the structure of a GG. In contrast with SVM's QP formulation that results in SVs, Chipclass is solely based on the graph formation rule and on distance calculation, not requiring user intervention for hyperparameter tuning nor optimization algorithms in order to be trained, being suitable for integrated-circuit implementation~\citep{Garcia_2020,janier2022}. The classifier results from a combination of the hyperplanes that bisect Support Edges (SEs), which are those graph edges that are formed by opposite-class vertices, i.e. $e_{ij}$ is a SE if $\mathbf{x}_i \in C_1$ and $\mathbf{x}_j \in C_2$ for a binary classification problem. The corresponding equations of all hyperplanes are obtained directly from the coordinates of $\mathbf{x}_i$ and $\mathbf{x}_j$, which are called Structural Support Vectors (SSVs) and are analogous to SVM's SVs. An example is given in Fig.~\ref{GGClass}, which shows a dataset of a binary classification problem with the corresponding GG and SEs. The final classifier is the one obtained by the aggregation of the straight lines that bisect each one of the three SEs, following Eq.~\ref{eq:prob_chipclass}.

\begin{equation}
	p(y=1|\textbf{x}) = \frac{w_p}{w_p+w_n}
	\label{eq:prob_chipclass}
\end{equation}

\begin{equation}
	w_p =  \sum\limits_{k=1}^H c_p
	\quad\text{and}\quad
	w_n =  \sum\limits_{k=1}^H c_n
	\label{eq:definition_gating}
\end{equation}

For each hyperplane $k$ and its corresponding SSVs from positive and negative classes, $\zeta_{kp}$ and $\zeta_{kn}$, Eq.~\ref{eq:definition_cpcn} implies that the class of the SSV closest to the test sample defines the contribution of that neuron to the final classification.

\begin{equation}
	(c_p, c_n) =\begin{cases}
		(c_k, 0), & \text{if } \delta (\textbf{x}, \zeta_{kp}) < \delta (\textbf{x}, \zeta_{kn}) \\
		(0, c_k), & \text{otherwise}
	\end{cases}
	\label{eq:definition_cpcn}
\end{equation}

\noindent $c_k$ weighs each neuron $k$ depending on its hyperplane distance to the test sample, as given by Eq.~\ref{eq:gating}, where $\textbf{p}$ is the vector of midpoints associated with each hyperplane.

\begin{equation}
	c_k = e^{\left(\frac{max(\delta(\textbf{x}, \textbf{p}_i))^2}{\delta(\textbf{x}, \textbf{p}_k)}\right)} \; \forall \; i= 1,...,H
	\label{eq:gating}
\end{equation}

\begin{figure}[htbp]
	\centerline{\includegraphics[scale=0.3]{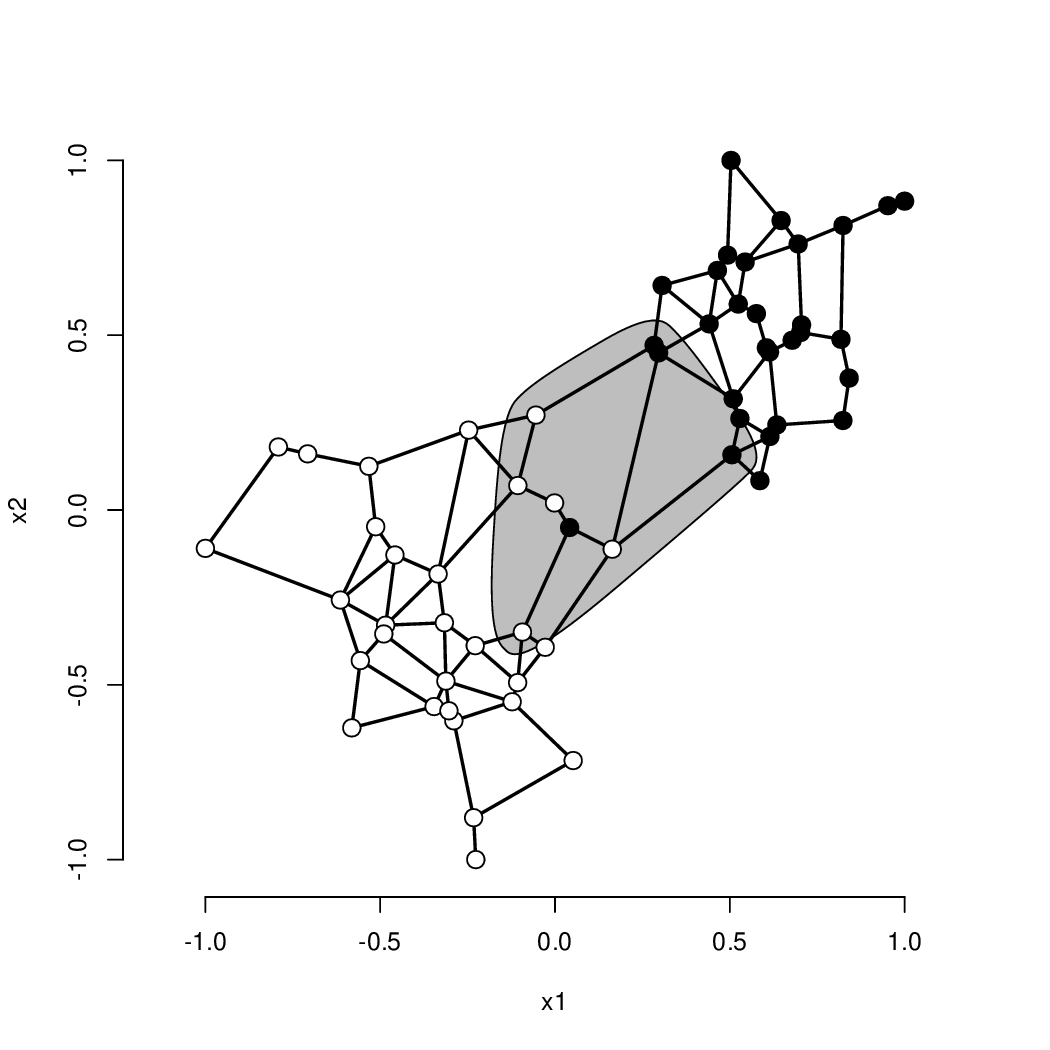}}
	\caption{GG for a binary classification dataset with class overlapping: an isolated sample from the black class generates additional SEs that yield an overfitted separation surface}
	\label{GGClass2}
\end{figure}

The example of Fig.~\ref{GGClass} showed a dataset without overlapping among samples from different classes. In most practical problems, however, superposition will occur and smoothing of the resulting separation surface should be accomplished in order to avoid overfitting. An example of such a situation is presented in Fig.~\ref{GGClass2}, where an isolated sample from the black class was observed in the vicinity of black samples. This isolated sample generates three additional SEs, one for each of its neighbors in the graph, yielding a separation surface contour that results in its classification in the black class. This overfitting effect happens because this isolated sample was considered as a SSV by construction, since all edges formed by opposite classes  determine a SE and two SSVs. This error minimization induction principle should be relaxed in order to smooth the surface and to reduce overfitting effects in this region where the black class is under-represented. This problem has been formulated in  SVMs by considering slack variables in order to relax the rigidity of the error constraint~\citep{Vapnik_1992} and requires that a regularization hyperparameter is provided in advance.

A filtering approach, which considers graph adjacency, was adopted in the  description of Chipclass and other GG-based classifiers. For instance, according to the original methodology such an isolated sample of Fig.~\ref{GGClass2} would be considered as an outlier and  discarded, since all its neighbors in the graph structure belong to the opposite class. Filtering will be discussed in more detail in the next section.

\subsection{Overlapping Filter Parameters}
One of the basic principles of Chipclass is to construct the classifier based solely on graph properties, making it simple to be implemented without the need to run computationally intensive optimization algorithms to induce the final model. Parsimonious models that trade-off optimality and computational cost and yet yield good performance can be obtained directly from the graph structure. This is the principle of the filtering procedure that considers the degree of a vertex and class labels in the adjacency in order to define a quality index for all training samples. The quality index is described next as presented in the original works~\citep{Aupetit_2005} and~\citep{Torres_2015}. Consider firstly the following definitions: $V_{dt}(\mathbf{x}_i)$ represents the degree of a vertex, defined as the total number of graph neighbors of $\mathbf{x}_i$, while $V{eq}(\mathbf{x}_i)$ denotes the number of graph neighbors belonging to the same class as $\mathbf{x}_i$. The quality index of a pattern $\mathbf{x}_i$ is defined according to Eq.~\ref{eqQx}.

\begin{equation}
	Q(\mathbf{x}_i) = \frac{ V_{eq}(\mathbf{x}_i)}{V_{dt}(\mathbf{x}_i)}
	\label{eqQx}
\end{equation}

\begin{figure}[h]
	\centerline{\includegraphics[scale=0.25]{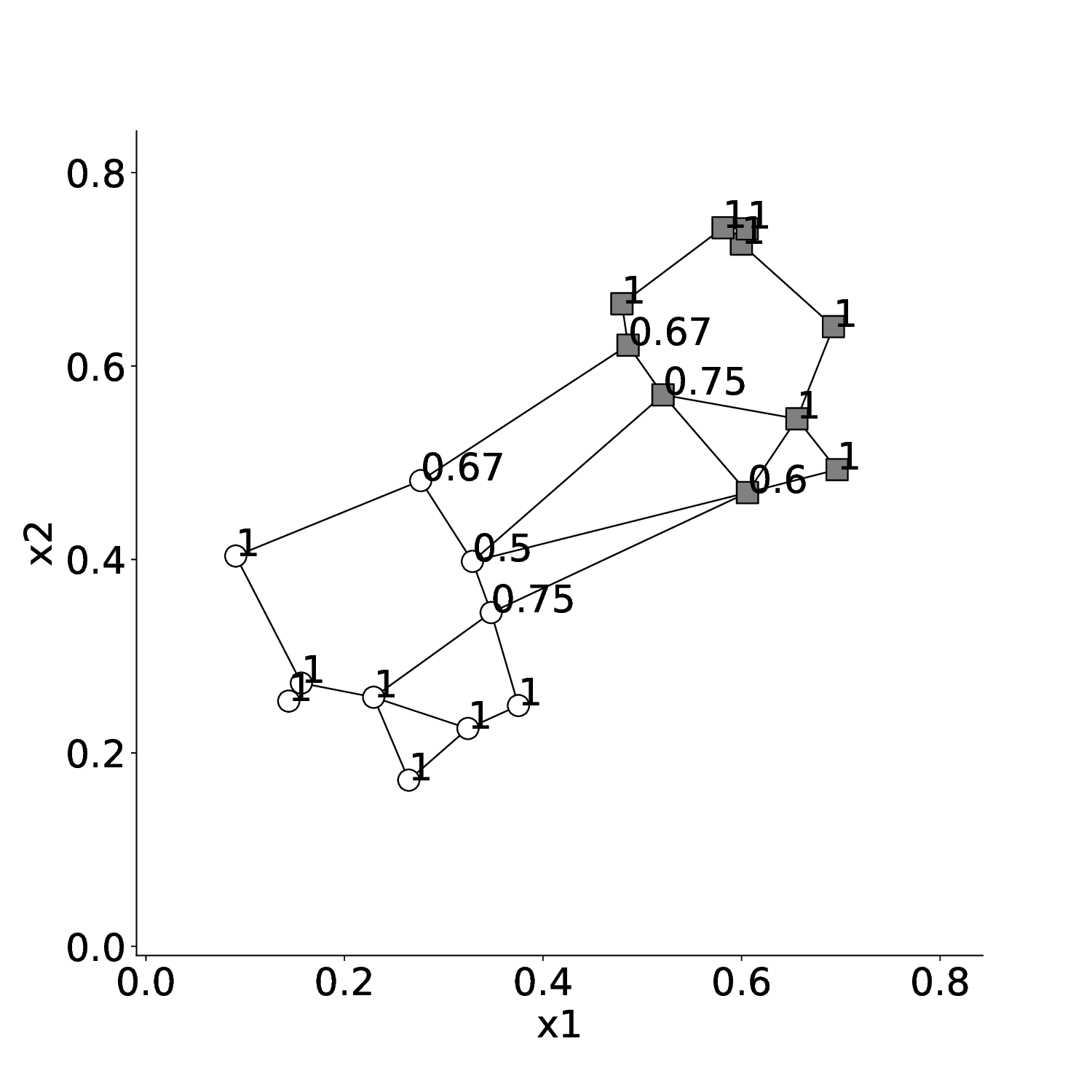}}
	\caption{$Q(\mathbf{x}_i)$ values for all samples from a binary classification problem. 6 samples on the margin have $Q(\mathbf{x}_i) < 1$, whilst the others that only have neighbors from the same class have $Q(\mathbf{x}_i) = 1$}
	\label{Q_metric_ex}
\end{figure}

\begin{figure}[h]
	\centerline{\includegraphics[scale=0.25]{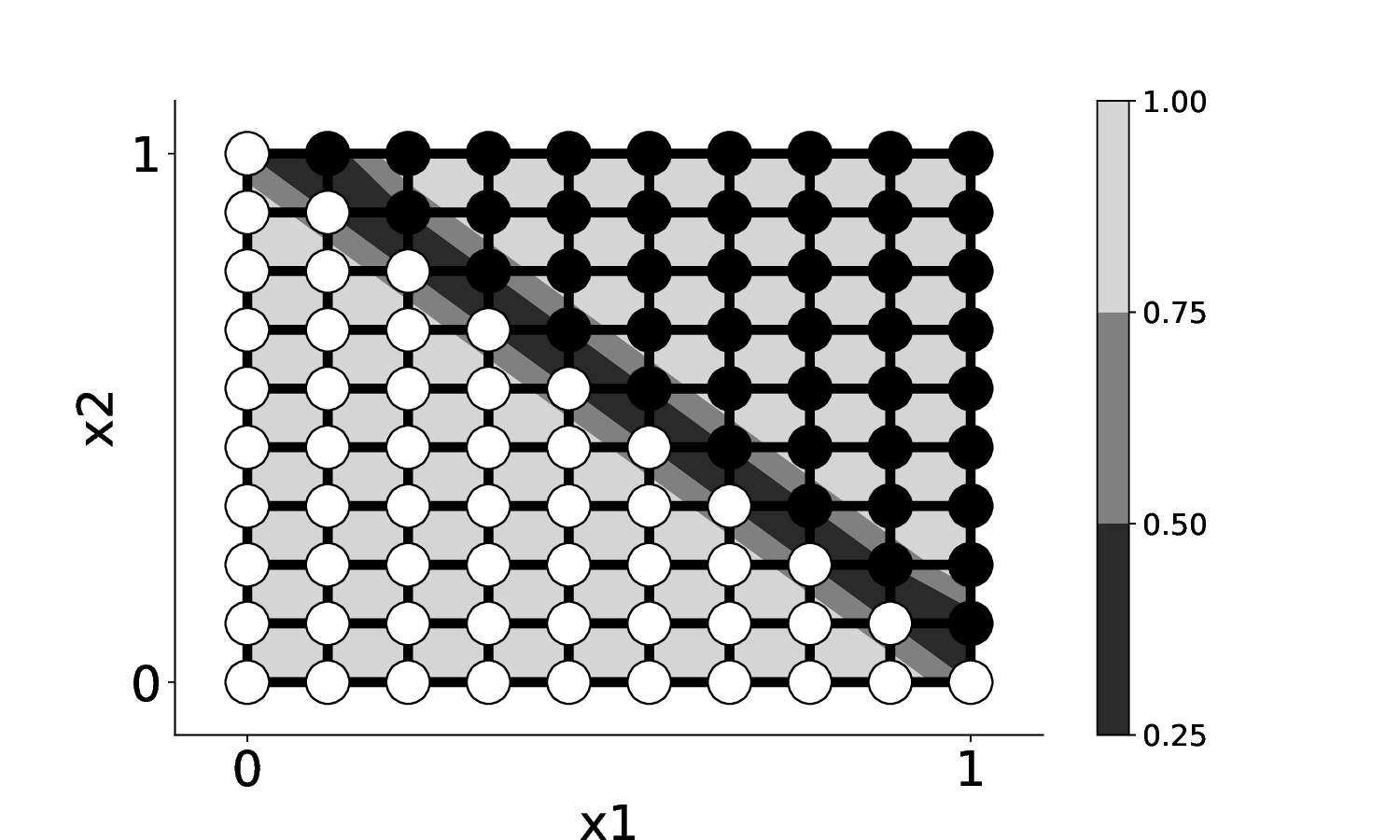}}
	\caption{$Q(\mathbf{x}_i)$ filled contours for a binary classification 2D grid: lower values are on the margin}
	\label{Qmetric_grid}
\end{figure}

Since $Q(\mathbf{x}_i)$ considers graph structure, the number of neighbors is known and spatial relations and densities can be assessed. For instance, $Q(\mathbf{x}_i)$ is analogous to the Lagrange multipliers that result from SVM's optimization since it provides information about pattern location in relation to the separation margin. Those samples with $Q(\mathbf{x}_i) = 0$, like the one in Fig.~\ref{GGClass2}, are likely to be outliers, while those with $Q(\mathbf{x}_i) = 1$ are likely to be located far from the separation margin. Those samples with $0 < Q(\mathbf{x}_i) < 1$ which are not outliers are located in the separation region. Outliers are discarded according to a threshold limit of $Q(\mathbf{x}_i)$ for each class. An example of $Q(\mathbf{x}_i)$ for a binary classification problem is presented in Fig.~\ref{Q_metric_ex}, where $Q(\mathbf{x}_i)$ values are shown for each sample. As it can be observed, smaller values occur for those samples  that are in the margin region, which is highlighted for a binary classification 2D grid shown in Fig.~\ref{Qmetric_grid}.

The threshold limit for removing outliers has been defined in the original work~\citep{Torres_2015} as the mean values $\theta_{+}$ and $\theta_{-}$ of $Q(\mathbf{x}_i)$ for each class, considering a binary classification problem with positive (+) and negative (-) samples. So, every sample with $Q(\mathbf{x}_i)<\theta_{+}\; \forall \mathbf{x}_i \in C_{+}$ and $Q(\mathbf{x}_i)<\theta_{-}\; \forall \mathbf{x}_i \in C_{-}$ is removed and the graph is recalculated without them. Previous works have shown that such a fixed and pre-established parameter results in parsimonious models that are competitive in performance with SVMs~\citep{Torres_2015}. This yields autonomy to the model, since it does not need to be adapted during training. As a trade-off between autonomy and performance, this paper aims at presenting an adaptive procedure for adjusting the threshold as a graph learning approach. The relationship between outlier filtering in the graph with regularization and margin maximization will be discussed in the next section.

\section{Margin Maximization and Graph-based Dataset Structure}
\label{sec:meth}

Chipclass implicitly maximizes the margin between SSVs, since
it is based on the bisecting hyperplane of the
edge that connects them. So, by construction, margin
is locally maximized for every pair of SSVs. Global
margin yielded by the aggregation of local classifiers can be estimated by
taking, for instance, the mean value of local margins. Another way to estimate the global margin in the absence of a hyperplane is to take the average of $M_i$ values calculated according to Eq.~\ref{eq:margin}~\citep{brahma2015deep}, where $\mathcal{M}$ is the nearest miss and $\mathcal{H}$ the nearest hit and, for the particular case of Chipclass,  $\textbf{x}_i \in SSVs$. $\mathcal{H}$ and $\mathcal{M}$ refer to samples from the opposite and from the same class of $\mathbf{x}_i$, respectively.

\begin{equation}
	M_i = \frac{||\textbf{x}_i-\mathcal{M}(\textbf{x}_i)||-||\textbf{x}_i-\mathcal{H}(\textbf{x}_i)||}{||\textbf{x}_i-\mathcal{M}(\textbf{x}_i)||}
	\label{eq:margin}
\end{equation}

Although the usual margin calculation is accomplished by
considering distances to hyperplanes, the outcome
obtained by the application of Eq.~\ref{eq:margin} provides an estimation of
the global margin. For instance, $M_i$ tends to be
negative for misclassified samples and positive for those that are correctly classified. The average of $M_i$ is taken on the whole dataset, however,
structural information provided by graph
relations, like the ones in Chipclass, may offer
a different perspective for the application of such a
measure. In other words, instead of applying
Eq.~\ref{eq:margin} to the whole dataset,
calculation can be directed to those points that
are known to provide a maximum margin local hyperplane between
classes, i.e., SSVs. Prior to filtering, every sample for which
$Q(\mathbf{x}_i) < \theta_{Class(\mathbf{x}_i)}$ is more likely to
contribute with negative terms for the summation of $M_i$, so the
elimination of such samples tends to increase the global margin of the aggregated separator.

\begin{figure}[h]
	\floatsetup[subfigure]{captionskip=0pt}
	\ffigbox{
		\begin{subfloatrow}[2]
			\ffigbox[\FBwidth]{\caption{} \label{fig:avg_margin}}{\includegraphics[width=\linewidth]{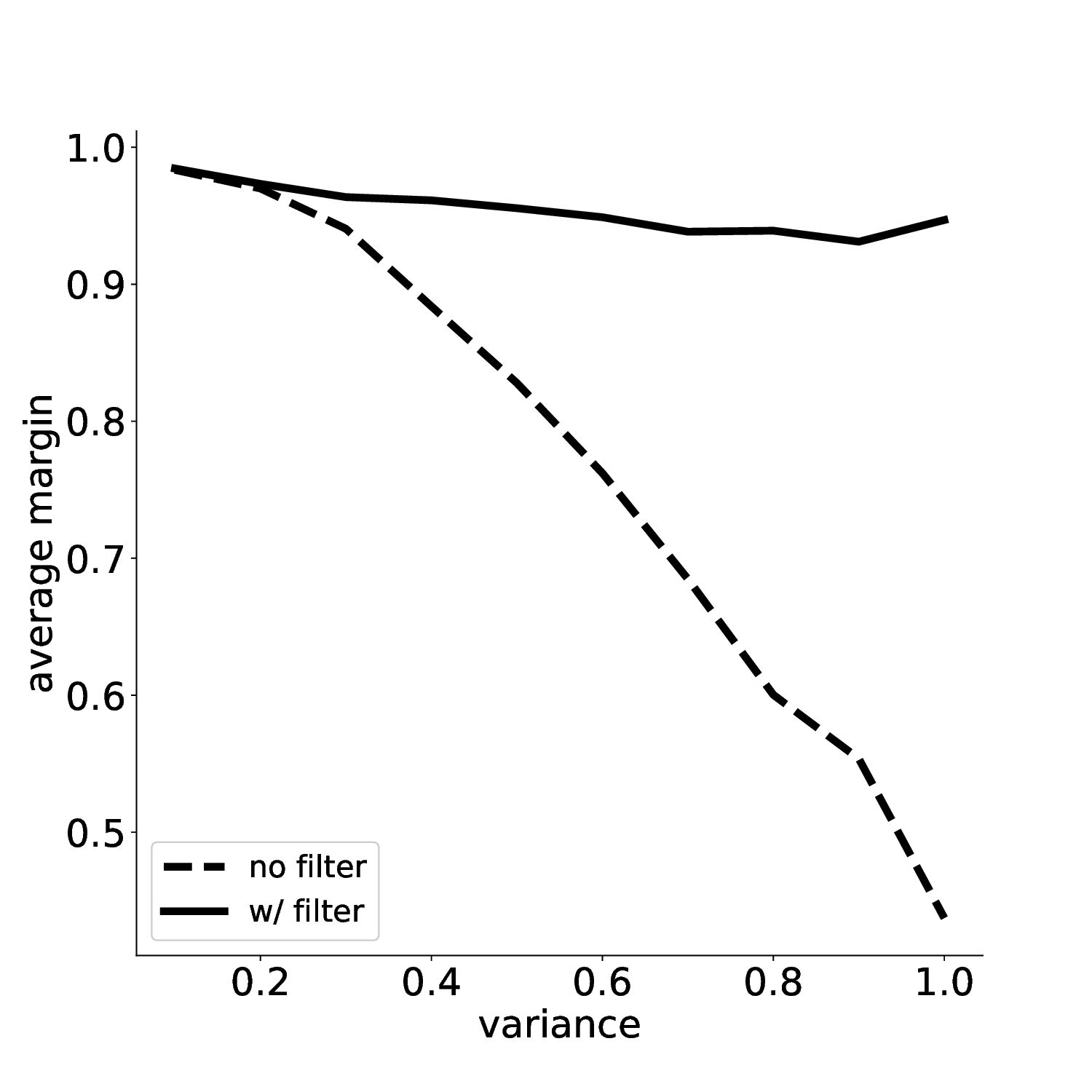}}
			\ffigbox[\FBwidth]{\caption{} \label{fig:avg_threshold}} {\includegraphics[width=\linewidth]{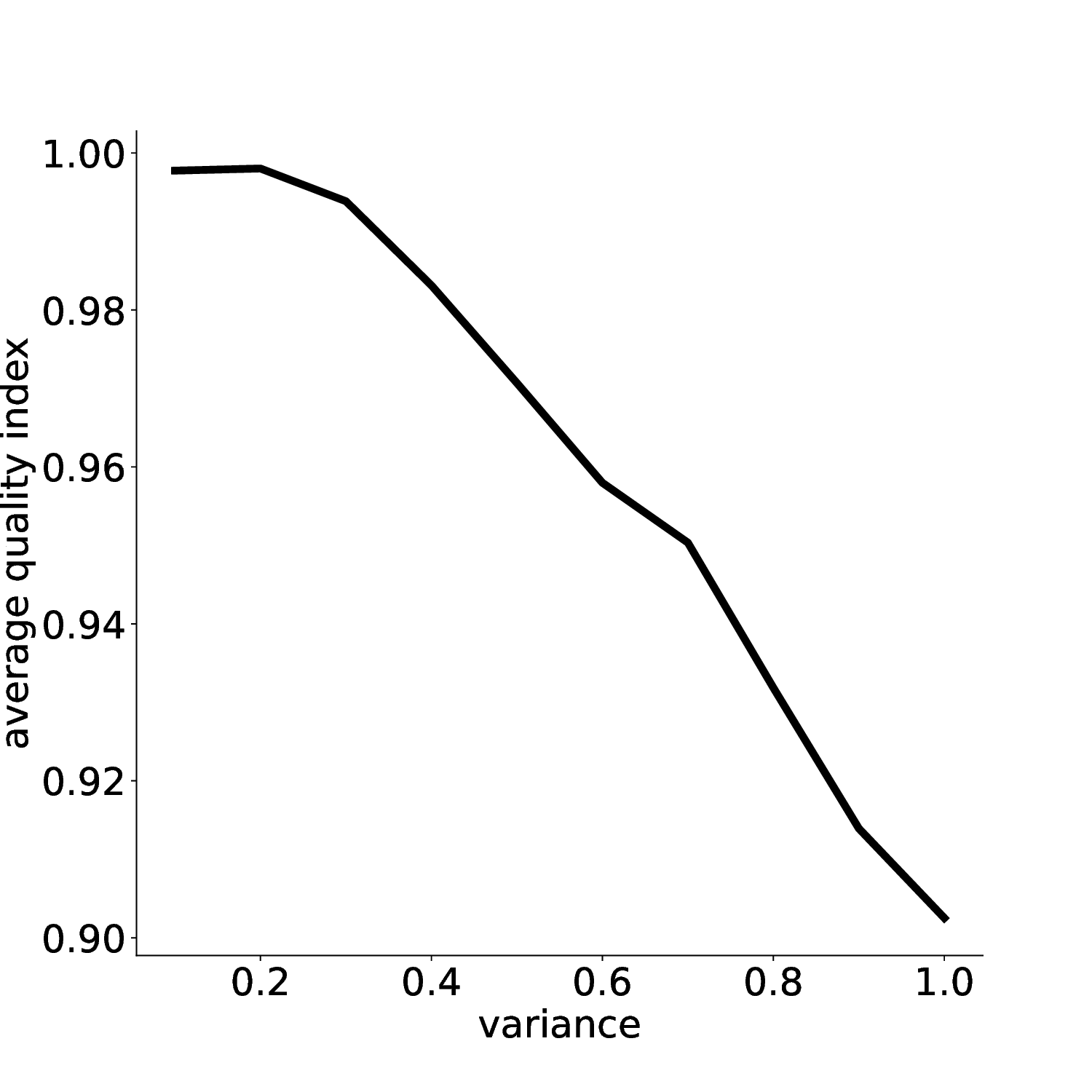}}
		\end{subfloatrow}
	}{\caption{Average margin (a) and quality index (b) for 2D-gaussian distributions with marginal variances varying between 0 and 1. (a) Filtering low quality index samples prevents the margin value from decaying with the increase of class overlapping, which occurs when there is no filtering. (b) As class overlapping increases, more samples tend to have neighbors from a different class, resulting in lower values of quality indexes.}
		\label{fig:avg_margin_threshold}}
\end{figure}

In order to show the relationship between the margin and the quality index $Q(\mathbf{x}_i)$, 2D-gaussian distributions with mean vectors $\mu_0 = (3,3)$ and $\mu_1 = (5,5)$, representing a binary classification problem, were generated for different covariance matrices, with null correlation coefficients and marginal variances varying from 0 to 1. For each variance value, the margin (Eq.~\ref{eq:margin}) and the quality index (Eq.~\ref{eqQx}) were computed for each sample and their means $\overline{M}$ and $\overline{\mathcal{Q}}$ were calculated. The resulting graphs of $\overline{M}$ and $\overline{\mathcal{Q}}$ as a function of the marginal variances are shown in Figs.~\ref{fig:avg_margin} and~\ref{fig:avg_threshold}. Since the means of the two generator distributions are fixed, it can be observed in the figures that the increase in variances leads to a corresponding decrease in both $\overline{M}$ and $\overline{\mathcal{Q}}$, due to the increase in the  overlap between the two classes. The graph of Fig.~\ref{fig:avg_margin} shows also $\overline{M}$ after removing all samples for which  $Q(\mathbf{x}_i)<\theta_{+}\; \forall \mathbf{x}_i \in C_{+}$ and $Q(\mathbf{x}_i)<\theta_{-}\; \forall \mathbf{x}_i \in C_{-}$. As it can be observed, the removal of samples results in a compensation of margin reduction due to overlapping that is observed in the experiment, which has an effect analogous to regularization yielded by slack variables in SVM's formulation.

\begin{table*}[b!t]
	\centering
	\scalebox{0.6}{
		\begin{tabular}{c|cccc|ccccccccc}
			\toprule
			\hline
			Dataset                                 & $d$ & $m$  & $m_{c+}$ & $m_{c-}$ & kNN     & Random Forest & SVM     & Chipclass & Chipclass flex. & GMM-GG  & GMM-GG flex. & RBF-GG  & RBF-GG flex. \\
			\midrule
			Abalone 18 vs. 9                        & 10  & 731  & 42       & 689      & 72.0831 & 82.935        & 90.2771 & 73.847    & 79.6939         & 71.127  & 74.3363      & 84.6718 & 89.5047      \\
			Appendicitis                            & 7   & 106  & 21       & 85       & 84.2014 & 84.6875       & 78.75   & 81.3194   & 79.0278         & 88.5417 & 88.4722      & 85.2778 & 85.4167      \\
			Australian                              & 14  & 690  & 307      & 383      & 91.8252 & 93.9249       & 92.3311 & 91.1915   & 91.2603         & 90.9019 & 92.0923      & 87.9006 & 92.4083      \\
			Banknote                                & 4   & 1348 & 610      & 738      & 99.8649 & 99.9889       & 100.0   & 99.8089   & 99.8909         & 100.0   & 100.0        & 100.0   & 100.0        \\
			Breast Cancer Wisconsin (Original)      & 9   & 449  & 236      & 213      & 98.3471 & 98.0758       & 98.7527 & 93.2853   & 94.9803         & 98.9511 & 99.0304      & 98.4551 & 98.3764      \\
			Breast Cancer Wisconsin (Prognostic)    & 32  & 194  & 148      & 46       & 58.4214 & 55.1048       & 61.7524 & 61.1762   & 57.0571         & 64.5095 & 64.0476      & 53.4333 & 60.3476      \\
			Climate Model  Simulation Crashes       & 18  & 540  & 494      & 46       & 89.6439 & 92.401        & 96.0245 & 94.0      & 94.6949         & 82.2378 & 88.7663      & 92.0439 & 93.1827      \\
			Fertility                               & 9   & 98   & 87       & 11       & 68.8542 & 77.5          & 66.8056 & 65.8333   & 64.5139         & 72.3611 & 64.8611      & 72.0833 & 57.8472      \\
			Glass Identification  7 vs. all         & 9   & 213  & 29       & 184      & 91.7105 & 96.8811       & 96.3353 & 95.2827   & 94.7271         & 96.6959 & 97.0663      & 96.3255 & 97.2904      \\
			Haberman's Survival                     & 3   & 277  & 204      & 73       & 59.7902 & 67.1526       & 70.0791 & 69.3457   & 70.1905         & 64.9069 & 60.0383      & 61.9847 & 67.8703      \\
			Statlog (Heart)                         & 13  & 270  & 150      & 120      & 87.4167 & 90.0556       & 89.3333 & 87.2222   & 88.1111         & 89.1111 & 89.9444      & 86.1667 & 89.7222      \\
			ILPD (Indian Liver Patient Dataset)     & 10  & 566  & 404      & 162      & 64.3709 & 73.8595       & 66.1996 & 67.3311   & 66.2296         & 58.645  & 69.5539      & 63.3411 & 66.6352      \\
			Ionosphere                              & 34  & 350  & 225      & 125      & 93.9875 & 97.6295       & 97.4807 & 93.9024   & 93.9538         & 92.0607 & 93.5895      & 98.7623 & 97.6708      \\
			Parkinsons                              & 22  & 195  & 147      & 48       & 97.4952 & 97.4524       & 97.7857 & 73.5762   & 89.9333         & 86.5619 & 97.8905      & 95.2143 & 97.1286      \\
			Statlog (Vehicle Silhouettes) 4 vs. all & 18  & 846  & 199      & 647      & 98.0914 & 99.4538       & 99.8491 & 94.7546   & 97.3592         & 96.0412 & 97.1487      & 99.8981 & 99.4295      \\
			Yeast 5 vs. all                         & 8   & 1453 & 51       & 1402     & 86.3159 & 91.3997       & 81.9772 & 89.4807   & 89.1783         & 80.1099 & 84.4121      & 85.475  & 87.7004      \\
			Yeast 9 vs. 1                           & 8   & 458  & 20       & 438      & 82.3692 & 89.2521       & 76.7627 & 74.8916   & 74.6644         & 80.6527 & 77.278       & 82.315  & 83.1105      \\
			\hline
			Avg. rank                               &     &      &          &          & 6.0588  & 3.4706        & 4.0     & 6.3529    & 6.0             & 5.7647  & 4.4706       & 5.2353  & 3.6471       \\
			\hline
			\bottomrule
		\end{tabular}
	}
	\caption{Average AUC for 10 Test Folds; Average Ranks; and Dataset Characteristics}
	\label{detail_table}
\end{table*}
\section{Conclusions and discussions}
\label{sec:conc}
\subsection{Relationship between margin and quality index}

The margin behavior observed in Figs.~\ref{fig:avg_margin} and \ref{fig:avg_threshold} can be best understood by expanding the mean of Eq.~\ref{eq:margin} in two summation terms, as in Eq.~\ref{eq:meanmargin}. The first term corresponds to those samples that will not be removed, with $Q(\mathbf{x}_i)$ above the threshold, while the second one corresponds to samples with $Q(\mathbf{x}_i)$ below the threshold and will be removed. Since the latter term has a larger contribution to the negative portion of the summation, there is a corresponding increase in the margin after their removal.

\begin{equation}
	\begin{split}
		\overline{M} = \frac{1}{m_{in}+m_{out}} \Big(\overbrace{\sum_{i=1}^{m_{in}}  \frac{||\textbf{x}_i-\mathcal{M}(\textbf{x}_i)||-||\textbf{x}_i-\mathcal{H}(\textbf{x}_i)||}{||\textbf{x}_i-\mathcal{M}(\textbf{x}_i)||}}^{Q(\mathbf{x}_i) \ge \theta_{Class(\textbf{x}_i)}}	+ \\
		\overbrace{\sum_{i=m_{in}+1}^{m_{out}} \frac{||\textbf{x}_i-\mathcal{M}(\textbf{x}_i)||-||\textbf{x}_i-\mathcal{H}(\textbf{x}_i)||}{||\textbf{x}_i-\mathcal{M}(\textbf{x}_i)||}}^{Q(\mathbf{x}_i) < \theta_{Class(\textbf{x}_i)}}\Big)
	\end{split}
	\label{eq:meanmargin}
\end{equation}

\noindent where $m_{in}$ is the number of samples that will be maintained and  $m_{out}$ is the number of samples that will be removed.

Similarly to Eq.~\ref{eq:margin}, the quality index of Eq.~\ref{eqQx} also provides margin information by considering the degree of a graph
vertex. $Q(\textbf{x}_i)$ can be interpreted as the normalized
within-class degree of a vertex, a property that is
intrinsic to the graph when class labels are
known. A separable situation occurs when
$Q(\mathbf{x}_i)=1 \; \forall \mathbf{x}_i \notin SSV$ and

\begin{equation}
	Q(\textbf{x}_i)=\frac{V_{dt}(\textbf{x}_i) - m_d}{V_{dt}(\textbf{x}_i)} \; \forall \mathbf{x}_i \in SSV,
\end{equation}

\noindent where $m_d$ is the number of SSVs directly connected to $\textbf{x}_i$. Thus, for this situation, heterogeneity in the neighborhood occurs only for $\textbf{x}_i \in SSV$, which have only its SSVs pairs from the opposite class.

In the situation when there is class overlapping, heterogeneity in the adjacency increases for those patterns that are not SSVs, what may lead to overfitting effects
due to all hyperplanes being considered to compose the final classifier. This is due to the error minimization principle of inductive learning and it is an intrinsic effect of every method that considers training error as an objective function. Most methods adopt regularization~\citep{Assis,SILVESTRE2015288} or structural shrinking~\citep{electronics9050811} in order to limit model capacity and to reduce error minimization effects in under-represented regions of the input space. SVM's formulation, for instance, considers slack variables to allow an additional degree of freedom in local error minimization. Lagrange multipliers, which result from such an approach, are responsible for weighting the importance of each sample in the final response of the model.

The threshold adopted in Chipclass has a similar effect of the slack variables and Lagrange multipliers of SVMs, since sample removal also results in smoothing the separation function in that region. The fixed threshold, according to the distribution of $Q(\mathbf{x}_i)$ for each class, complies with the Chipclass principle of not relying on extensive optimization methods, being user-independent and feasible to be implemented in hardware. This paper explores the idea that threshold could still be adapted according to the problem.

In order to provide an
additional degree of freedom to the classifier,
a variable threshold is considered in this paper, as
presented in Eq.~\ref{eq:filter_def}.

\begin{equation}
	h_{class(\textbf{x}_i)} \cdot Q(\textbf{x}_i) < \theta_{class(\textbf{x}_i)}
	\label{eq:filter_def}
\end{equation}

\noindent where $h_{class(\textbf{x}_i)}$ is a per-class hyperparameter that modulates the quality index $Q(\textbf{x}_i)$. Rather than using fixed thresholds for each class, as in $Q(\textbf{x}_i) < \theta_{class(\textbf{x}_i)}$, the hyperparameter $h_{class(\textbf{x}_i)}$ can be selected during a hyperparameter tuning process in the training phase. This allows for the exploration of the solution space and enables flexible threshold selection based on this optimization process. In the special case where $h_{class(\textbf{x}_i)} = 1$, the formulation reduces to the original fixed-threshold version.

\subsection{Regularization with sample removal}

Considering the framework of statistical learning~\citep{vapnik1995nature} and assuming equal costs to errors of different classes, the minimization of the expected risk (Eq.~\ref{eq:expected_risk}) for a finite set of input samples drawn from an unknown joint probability density function $p(\textbf{x},\textbf{y})$ is achieved when estimating the optimal function $f_0(\textbf{x})$ (Eq.~\ref{eq:optm_fun}) for a binary classification problem, where $\mathcal{R}_0$ and $\mathcal{R}_1$ are two disjoint regions separated by the decision surface denoted by $f_0(\textbf{x})$~\citep{berger2013statistical, duda2006pattern}.

\begin{equation}
	\begin{split}
		R[f] & = P(\textbf{x} \in \mathcal{R}_1, y=0) + P(\textbf{x} \in \mathcal{R}_0, y=1)                        \\
		     & = \int_{\mathcal{R}_1}p(\textbf{x},y=0)d\textbf{x}  \int_{\mathcal{R}_0}p(\textbf{x},y=1)d\textbf{x}
	\end{split}
	\label{eq:expected_risk}
\end{equation}

\begin{equation}
	f_0(\textbf{x})=\begin{cases}
		1, & \text{if } \frac{p(\textbf{x}|y=1)}{p(\textbf{x}|y=0)} \geq \frac{P(y=0)}{P(y=1)} \\
		0, & \text{otherwise}
	\end{cases}
	\label{eq:optm_fun}
\end{equation}

Thus, optimization of supervised learning problems under such framework includes finding the likelihood $p(\textbf{x}|y=k)$ and the prior probabilities for each class $k$.

For Chipclass, the decision surface is given by Eq.~\ref{eq:decision_Chipclass}, where $H$ is the number of boundary hyperplanes defined by the GG.

\begin{equation}
	\hat{f}(\textbf{x})=\begin{cases}
		1, & \text{if } \frac{w_p}{w_n} \geq 1 \\
		0, & \text{otherwise}
	\end{cases}
	\label{eq:decision_Chipclass}
\end{equation}

\begin{figure}[h]
	\floatsetup[subfigure]{captionskip=0pt}
	\ffigbox{
		\begin{subfloatrow}[2]
			\ffigbox[\FBwidth]{\caption{} \label{fig:c_no_filter}}{\includegraphics[width=\linewidth]{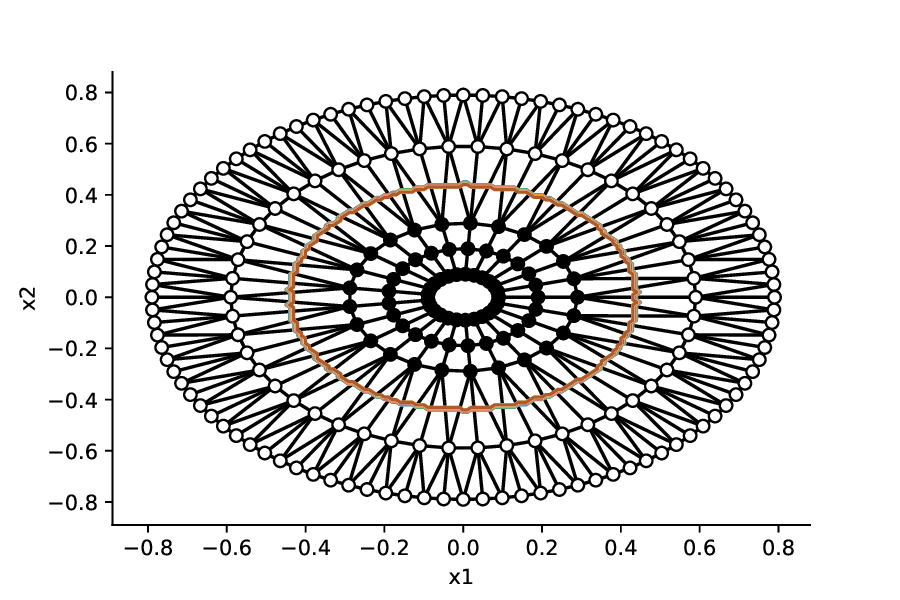}}
			\ffigbox{\caption{} \label{fig:c_filter_Chipclass}} {\includegraphics[width=\linewidth]{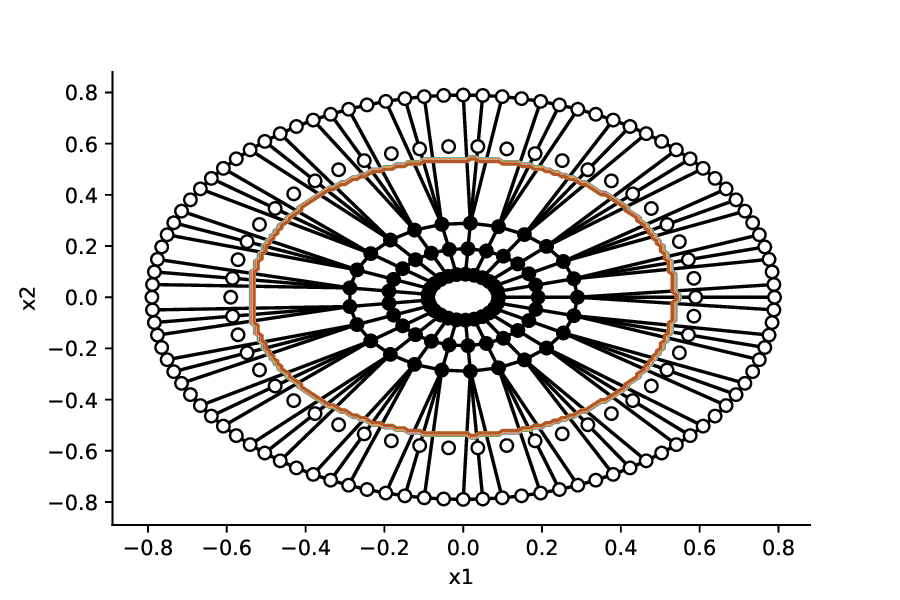}}
		\end{subfloatrow}
	}{\caption{Chipclass' separation surface for a binary classification problem when: (a) regularization is not applied and therefore all samples in the graph are used. (b) regularization is applied: while the white class has the margin samples filtered out, the imbalanced black class receives a higher $h_{class(\textbf{x}_i)}$ and ends up not having any filtered samples, which expands the classification surface in its favor}
		\vspace*{-3mm}
		\label{fig:c_filter_nofilter_Chipclass}}
\end{figure}

Thus, decision surface estimation depends upon the distance between the test sample $\textbf{x}$ and the SSVs of each hyperplane. When considering class overlapping, the algorithm not only estimates new hyperplanes, but most importantly yields new distance relationships between new SVs and the test sample. These relationships directly affect the decision surface ratio presented in Eq.~\ref{eq:decision_Chipclass}, thus, adding a scale factor $h_{class(\textbf{x}_i)}$ to the threshold definition of each class allows for changing the prior probability of the class being evaluated, as different SSVs will be considered. Figs.~\ref{fig:c_no_filter} and~\ref{fig:c_filter_Chipclass} show two decision surfaces, for Chipclass without regularization and Chipclass considering the pre-established threshold for white samples and an $h_{class(\textbf{x}_i)}$ that allows all black samples to be considered in the classifier definition. As it can be seen, such flexibility affects the decision surface mapping of the classifier.

\begin{figure}[h]
	\centerline{\includegraphics[scale=0.2]{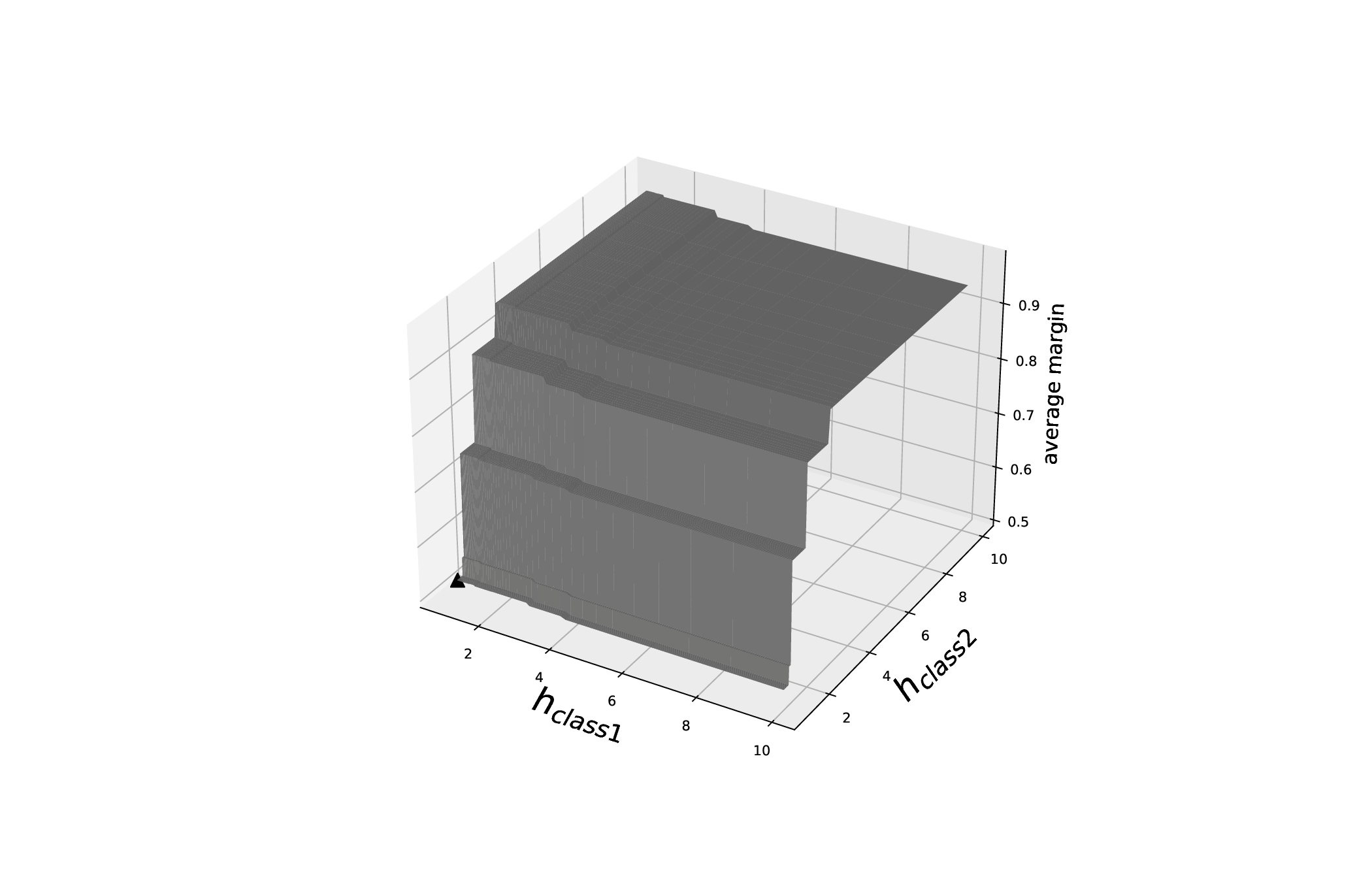}}
	\caption{Average margin values as a function of $h_{class 1}$ and $h_{class 2}$}
	\vspace*{-10mm}
	\label{fig:margin_variation}
\end{figure}

\begin{figure*}[t]
	\floatsetup[subfigure]{captionskip=0pt}
	\ffigbox{
		\begin{subfloatrow}[3]
			\ffigbox[\FBwidth]{\caption{} \label{fig:no_filter}}{\includegraphics[width=\linewidth]{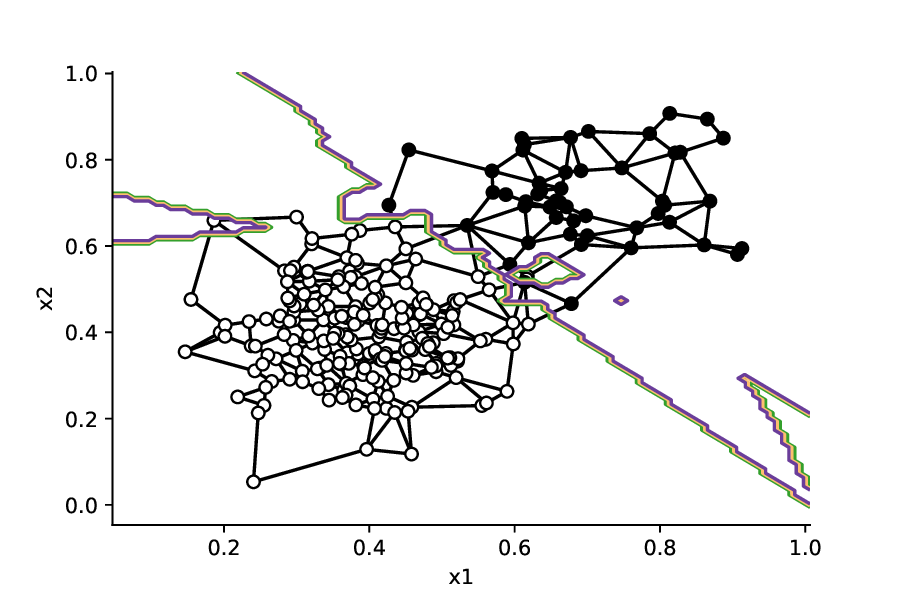}}
			\ffigbox[\FBwidth]{\caption{} \label{fig:filter_Chipclass}} {\includegraphics[width=\linewidth]{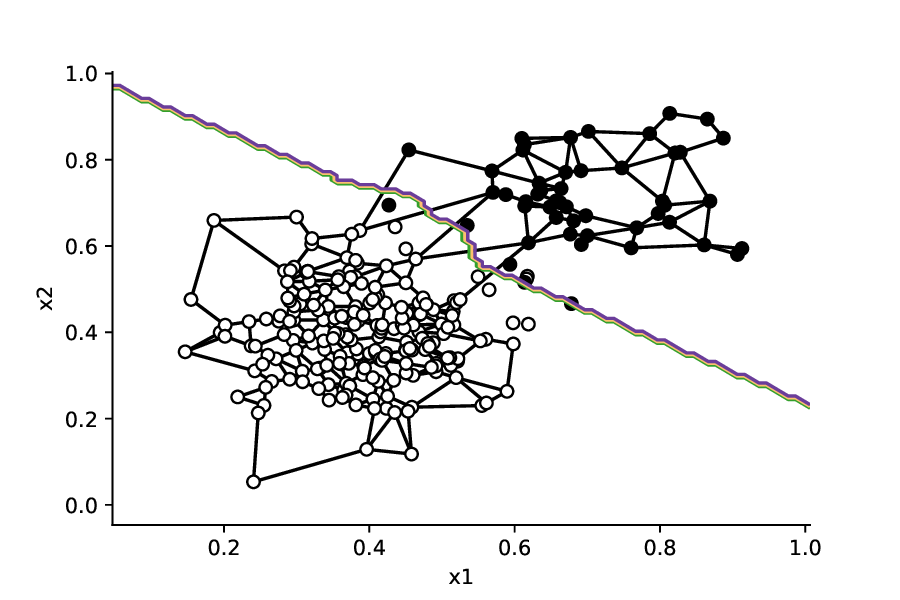}}
			\ffigbox[\FBwidth]{\caption{}\label{fig:filter_hyper}}{\includegraphics[width=\linewidth]{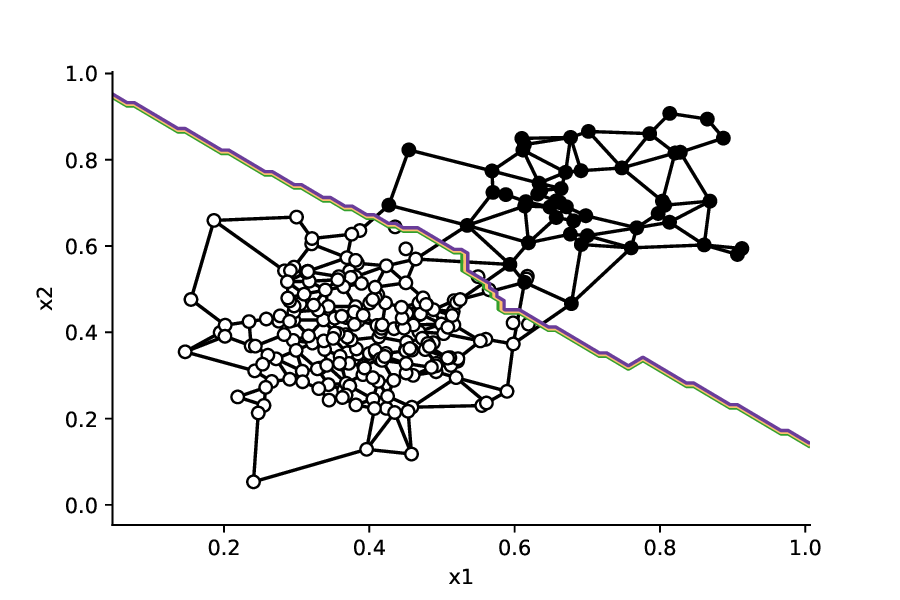}}
		\end{subfloatrow}
	}{\caption{Chipclass' separation surface when: (a) regularization is not applied and all samples are considered. (b) the fixed threshold ($h_{class(\textbf{x}_i)} = 1$) is considered. (c) the two classes have different $h_{class(\textbf{x}_i)}$ values. The black class has a lower threshold, which maintains its samples in the Graph and therefore shifts the separation surface to the opposite direction.}
		\label{fig:nofilter_cfilter_hfilter_Chipclass}}
\end{figure*}

Such flexibility may also address class imbalance, for which machine learning algorithms tend to prioritize the majority class~\citep{castro2013novel}. High values of $h_{class(\textbf{x}_i)}$ for the minority class and low values for the majority class may provide results similar to techniques such as oversampling/undersampling the minority/majority class or using different loss functions for different classes~\citep{castro2013novel}. Figs.~\ref{fig:no_filter},~\ref{fig:filter_Chipclass} and~\ref{fig:filter_hyper} illustrate such an effect, where low and high values of $h_{class(\textbf{x}_i)}$ for the minority and majority classes, respectively, lead to a shift in the decision surface towards the disjoint region of the majority class.

Considering two 2D-gaussian distributions with 500 samples each, centered on $\mu_0 = (3,3)$ and $\mu_1 = (5,5)$ with covariance matrices with null correlation coefficients and marginal variance equal to 0.3, representing 2 classes of a binary classification problem, Fig.~\ref{fig:margin_variation} shows the average margin value, as a function of $h_{class1}$ and $h_{class2}$, for the samples with $Q(\textbf{x}_i) \geq \frac{\theta_{class(\textbf{x}_i)}}{h_{class(\textbf{x}_i)}}$. The higher $h_{class(\textbf{x}_i)}$ the lower the threshold and fewer samples are filtered. Whilst the fixed threshold formulation results in one point on the surface, shown by the triangle marker at $h_{class1} = 1$ and $h_{class2} = 1$, the per-class hyperparameter opens up a finite range of possibilities that yield different average margin values, as well as different support edges and metric scores.

\section{Experiments and Results}
\label{sec:exp}
Experiments were carried out with 15 datasets from the UCI repository~\citep{Dua_2019}, and one from the KEEL-dataset repository~\citep{Alcal_2011} (Appendicitis).

Standard Chipclass~\citep{Torres_2015}, GMM-GG~\citep{torres2020large} and RBF-GG~\citep{torres2014geometrical} (with fixed thresholds $\theta_+$ and $\theta_-$, so that the filtering approach follows $Q(\textbf{x}_i) < \theta_{class(\textbf{x}_i)}$, based on the methodology of~\citep{torres2020large}) were compared with their versions using the flexible threshold proposed in this paper (Chipclass flex., GMM-GG flex. and RBF-GG flex.), with tuned hyperparameters $h_{class1}$ and $h_{class2}$ of Eq.~\ref{eq:filter_def}, and literature models: SVMs, Random Forests and k-Nearest Neighbors (kNN). RBF-GG was implemented with the activation function proposed in~\citep{hanriot2024multiclass}.

Table~\ref{detail_table} presents the mean AUC (Area Under the ROC Curve) of 10 test folds: for each fold, the remaining samples were used to train the models and tune their hyperparameters in a 5-fold cross-validation. Hyperparameters of the flexible GG-based classifiers and literature models were tuned with Bayesian Optimization with 50 trials, where the objective function to be maximized was the mean AUC of the validation sets. Each classifier's average Friedman rank~\citep{Janez_2006_compar}, as well as core features of each benchmark dataset adopted in the present work were also presented. $m$ denotes number of samples, $d$ the number of features, and $m_{c+}$ and $m_{c-}$ the number of patterns labeled as (+1) and (-1), respectively.

Applying the Friedman test for the comparison of multiple classifiers~\citep{Janez_2006_compar}, $F_f = 3.22$ for 9 classifiers and 17 datasets, with F(8,128) at $\alpha=0.05$ being equal to 2.01. Since $F_f > F(8,128)$, the null hypothesis $H_0$ that the classifiers are equal can be rejected. Applying the Bonferroni-Dunn $post hoc$ test, and assuming $\alpha=0.05$, $q_a = 2.724$, thus $CD=2.5588$. GG-based classifiers with flexible thresholds present better average ranks than their standard versions and are within the critical value compared to random forests and SVMs.

Chipclass was originally described with the aim to provide a classifier that does not depend on user intervention to set parameters and also does not require a host processor to run optimization~\citep{Torres_2015}. GMM-GG~\citep{torres2020large} and RBF-GG~\citep{torres2014geometrical} also used a fixed filtering parameter and therefore did not go under any hyperparameter tuning process. Here, we exploit the regularization effect of GG-based models, adding a per-class hyperparameter that makes the filtering threshold flexible. We discuss how such addition impacts on the global margin value of the final classifier and enhances its performance, as flexible thresholds expand the solution space, which was in previous works a single solution defined by the averages of the quality indices of each class. We then applied a Friedman test to show that using flexible thresholds defined with hyperparameter tuning improves all standard GG-based classifiers versions.

\section{Acknowledgments}

The authors would like to thank the Brazilian agencies CAPES and CNPq for the financial support.

\bibliographystyle{model2-names}

\bibliography{ref9}

\end{document}